\documentclass[letterpaper, 10pt, conference]{ieeeconf}
\IEEEoverridecommandlockouts    

\usepackage[pdftex]{graphicx}
\usepackage{amsmath}
\usepackage{algorithmic}
\usepackage{array}
\usepackage[caption=false,font=footnotesize,labelfont=sf,textfont=sf]{subfig}
\usepackage{url}
\usepackage{amsfonts}
\usepackage{amssymb}
\usepackage{bm}
\usepackage{adjustbox}
\usepackage{multicol}
\usepackage{tabularx}
\usepackage{makecell}
\usepackage{verbatim}
\usepackage{color}
\usepackage{xspace}
\usepackage{graphics}
\usepackage[normalem]{ulem}

\def\etal{\mbox{\textit{et al.}}\@\xspace}

\title{\LARGE \bf PerceptNet: Learning Perceptual Similarity of Haptic Textures in Presence of Unorderable Triplets}

\author{Priyadarshini K$^{1}$, Siddhartha Chaudhuri$^{2}$, and Subhasis Chaudhuri$^{3}$
\thanks{$^{1}$ IIT Bombay, India.
        {\tt\small priyadarshini.k@iitb.ac.in},}%
\thanks{$^{2}${Adobe Research and IIT Bombay.}
        {\tt\small sidch@cse.iitb.ac.in},}%
\thanks{$^{3}$ IIT Bombay, India.
        {\tt\small sc@ee.iitb.ac.in}}%
}

\begin{document}

\maketitle
\thispagestyle{empty}
\pagestyle{empty}

\begin{abstract}
In order to design haptic icons or build a haptic vocabulary, we require a set of easily distinguishable haptic signals to avoid perceptual ambiguity, which in turn requires a way to accurately estimate the perceptual (dis)similarity of such signals. In this work, we present a novel method to learn such a perceptual metric based on data from human studies. Our method is based on a deep neural network that projects signals to an embedding space where the natural Euclidean distance accurately models the degree of dissimilarity between two signals. The network is trained only on non-numerical comparisons of triplets of signals, using a novel triplet loss that considers both types of  triplets that are easy to order (inequality constraints), as well as those that are unorderable/ambiguous (equality constraints). Unlike prior MDS-based non-parametric approaches, our method can be trained on a partial set of comparisons and can embed new haptic signals without retraining the model from scratch. Extensive experimental evaluations show that our method is significantly more effective at modeling perceptual dissimilarity than alternatives.

\end{abstract}

\section{Introduction}

Tactile sensory inputs can be valuable for user-facing applications requiring {\em symbolic or cross-modal communication}. For instance, tactile/haptic feedback can notify us about events, provide information on sensory data that is out-of-spectrum or in an unavailable modality, and act as a communication medium for people with visual or auditory impairments~\cite{hapticApp1, hapticApp2}. All such applications require a mapping between haptic signals and the domain vocabulary, e.g. different colors, or different events, or different letters and words. In order to build any such application, therefore, it is critical to have a set of haptic signals which are easily distinguishable to avoid perceptual ambiguity. In turn, this requires that we have a robust {\em perceptual} metric that models the human-perceived dissimilarity of different haptic signals.

Developing such a metric is difficult. {\em First}, humans are bad at estimating the degree of (dis)similarity of different signals on a consistent scale \cite{GNMDS}. Thus, we must rely on coarse-grained boolean comparisons that only indicate whether a person can distinguish between two signals or not, possibly with reference to a base signal. {\em Second}, complex nuances of human perception such as just-noticeable-difference (JND)~\cite{JND_est} require metric learning to extract useful information not just from easily distinguishable exemplars, but also from {\em indistinguishable} exemplars, again possibly with reference to a base signal. {\em Third}, gathering large amounts of haptic training data is tedious and expensive. For instance, prior non-parametric metric learning methods such as the MDS-based approach of Hollins \etal~\cite{hollins1993} require a complete set of comparisons of all possible pairs of signal types.

In this work, we propose a novel parametric approach for learning a perceptual metric for haptic signals. Our method utilizes a deep neural network that learns to project low-level features of input signals to an embedding space where the Euclidean distance accurately reflects the human-perceived dissimilarity between signals. The network is trained with triplets of signals: each triplet indicates (a) if one signal is more similar to a base signal than another (an easily-ordered or {\em high-margin} triplet), {\em or} (b) if no such ordering can be reliably deduced (an unorderable or {\em low-margin} triplet). Such comparisons are easily obtained via human studies~\cite{Psychometric}, and are informative for metric learning {\em without} requiring numerical estimates of the distance between two signals~\cite{FaceNet}. Unlike prior triplet-based learning approaches which tend to ignore triplets of the second type as uninformative~\cite{mcfee2011, perceptualMetric}, our approach treats both triplet types (with, or without, a clear ordering) as informative: we show this improves the learned metric. Since our method is parametric, we can train it on as much, or as little, data as is available: it generalizes gracefully to situations where many combinations of signals lack comparisons, and applies even when some signal types are completely unseen during training.

This paper focuses on {\em haptic textures}: force feedback recordings of different surface materials which can be played back via a haptic renderer. We use the texture database of Strese \etal~\cite{TUM},
and use a (novel) spectral representation of each texture as the input to the neural network. However, our metric learning approach is generally applicable, and can be used for any collection of signals where basic input features and triplet comparisons are available.

Some prior works use neural networks for {\em semantic} separation, to cluster signals by class label \cite{zheng2016deep, gao2016deep, kerzel2017haptic}. Note that the underlying spirit of our work is significantly different from semantic separation tasks. In semantic separation, the relationship among input instances is binary: all signals from the same class are considered equally similar, and all signals from different classes are considered equally different. In contrast, human perception is more granular: two classes could be {\em more dissimilar} than two other classes. For instance, the surface textures of ``Brass'' and ``Rubber'' could be perceived as more distinct than two other textures ``Brass'' and ``Copper''. Hence, classical approaches of material classification are insufficient for finding a perceptual embedding of the signal space.

In summary, we propose the first deep metric learning approach for modeling the perceptual similarity of haptic signals. The model is trained only on non-numerical, relative comparisons of signals, with a novel loss function accommodating both triplets that can and cannot be ordered. The parametric framework can be trained on relatively little data, and generalizes to unseen signals. Extensive experiments show that our method improves upon prior alternatives.

\section{Related Work}

We overview the related work on semantic and perceptual embedding of signals of different modalities:

\paragraph{Semantic Embedding of Tactile Signals}
Semantic embedding has been well explored in the vision and speech domains, where objects are classified into categorical labels using machine learning methods \cite{img2013survey, audio2011survey}. However, classification of materials using tactile sensory input has only started recently with the emergence of haptic data recording tools \cite{kerzel2017haptic, zheng2016deep, gao2016deep, strese2015surface, liu2018material}. Gao \etal~\cite{gao2016deep} use both visual and haptic data to classify surface materials with haptic adjectives such as hard, metallic, fuzzy etc. Liu \etal~\cite{liu2018multi} proposed a dictionary learning model for material categorization using audio and acceleration data. Another interesting work by Aujeszky \etal~\cite{aujeszky2018material} shows improvement in classification results by considering physical properties of the ambient environment along with object properties. In contrast to our approach, all these methods are class-dependent and cannot model human-perceived similarity.

\paragraph{Perceptual Embedding of Tactile Signals}
Our work is inspired by Enriquez \etal~\cite{phonemes, maclean2003perceptual}, who develop a set of perceptually well-separated haptic icons using MDS (multi-dimensional-scaling). Other relevant MDS-based work includes \cite{hollins1993, pasquero2006, vocab_MDS}. Although this approach has proved useful for perceptual embedding, it has some limitations. {\em First}, it requires ground-truth comparisons of all possible pairs of signals: the quadratic scaling is poor for larger datasets.  {\em Second}, it requires numerical estimates of pairwise distances, which are hard for individual humans to provide and can be estimated only by statistical aggregation over large user studies. {\em Third}, MDS does not support uncertainty in relative orderings. {\em Fourth}, it is a non-parametric technique which does not apply to novel signals. In contrast, our method addresses all four limitations with a parametric solution, trained only on non-numerical triplet orderings, which can be directly applied to embed novel signals. Our framework can work with partial training data and incorporates uncertainty in relative ordering into the modeling process.

\paragraph{Perceptual Embedding of Visual and Audio Signals}
Several prior works in vision and speech processing proposed parametric models for perceptual metrics, using deep learning or classical kernel methods \cite{perceptualMetric, FCN, DAF, mcfee2011}. Our work has similar goals, albeit in the haptic domain. However, none of these prior works leverages unorderable/ambiguous triplets. Pei \etal~\cite{DNK} incorporate such constraints for image clustering. Their method assumes triplet constraints are based on an underlying set of latent class labels: each triplet has two samples from one class and a third from another class. We make no such assumption, since we require a perceptual and not a semantic embedding: our triplets typically compare signals from three different classes.

\section{Method}
\label{method}

We are given a set of haptic signals \mbox{$X=\{x_{i}\}_{1}^m \in \mathbb{R}^{n}$}, each described by $n$ features, and a set of constraints \mbox{$C = H \cup L$} encoding relative comparisons between the signals. Each constraint is a triplet of signals \mbox{$(x_{i}, x_{j}, x_{k})$} belonging to one of two types: {\em high-margin} ($H$) or {\em low-margin} ($L$). A {\bf high-margin triplet} indicates a case where humans are able to clearly identify signal $x_i$ as being more similar to $x_j$ than to $x_k$. In other words, it captures the inequality relation \mbox{$d^*(x_i, x_k) - d^*(x_i, x_j) \geq \xi^*$}, where $d^*(\cdot, \cdot)$ is the ground-truth perceptual distance (dissimilarity) between two signals, and \mbox{$\xi^* > 0$} is some minimum margin representing the threshold of human discrimination. Conversely, a {\bf low-margin triplet} indicates that humans are uncertain whether $x_j$ or $x_k$ is more similar to $x_i$, i.e. it encodes the (approximate) equality relation \mbox{$| d^*(x_i, x_k) - d^*(x_i, x_j) | < \xi^*$}. Triplet-based orderings are easy to crowdsource (Section \ref{experiments} describes our acquisition process), more reliable than numerical similarity judgements, and effective for training metric learning models~\cite{FaceNet, Psychometric}. The consideration of low-margin triplets in addition to high-margin triplets, accounting for human uncertainty in perceptual embedding, is a contribution of our method.

Our goal is to learn an embedding kernel \mbox{$\phi : \mathbb{R}^n \rightarrow \mathbb{R}^m$}, such that the Euclidean distance \mbox{$d_\phi(x, y) = \| \phi(x) - \phi(y) \|$} satisfies the triplet constraints as well as possible for a learned margin $\xi_\phi$. In other words, \mbox{$\forall c = (x_i, x_j, x_k) \in C$},
\begin{equation} \label{prob_def}
\begin{alignedat}{4}
  & d_\phi(x_i, x_k) - d_\phi(x_i, x_j) && ~\geq~ && \xi_\phi \qquad && \textrm{if~} c \in H \\
  & | d_\phi(x_i, x_k) - d_\phi(x_i, x_j) | && ~<~ && \xi_\phi \qquad && \textrm{if~} c \in L
\end{alignedat}
\end{equation}
Once learned, the metric $d_\phi$ can be used to estimate the perceptual dissimilarity of any pair of seen or unseen signals.

\begin{figure}
\centering
\includegraphics[scale=0.4]{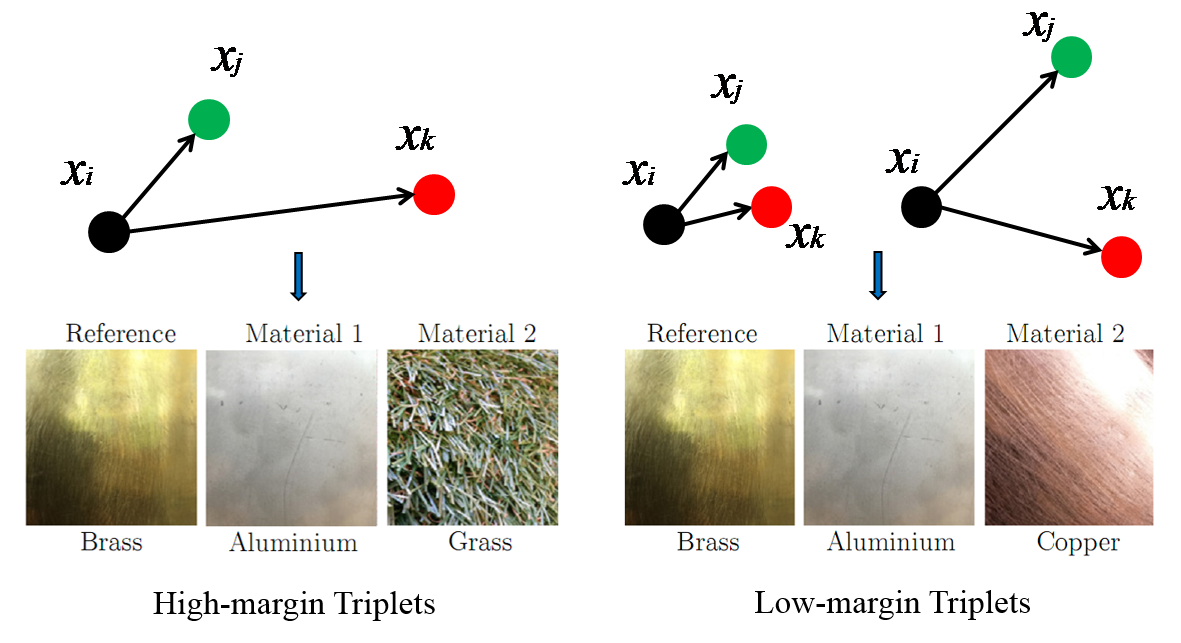}
\vspace{-1mm}
\caption{Examples of high- and low-margin triplets of haptic textures~\cite{TUM}. The first material in each triplet is the reference. The high-margin triplet {\em (left)} shows brass is perceptually closer (in terms of haptic feedback) to aluminum than grass; the low-margin triplet {\em (right)} shows brass is perceptually as similar to aluminum as is copper.}
\vspace{-3mm}
\label{triplet_type}
\end{figure}

\subsection{Learning the Perceptual Distance Metric}
\label{learning}

We use a deep neural network (DNN) to learn the embedding $\phi$ that defines our distance metric $d_\phi$. DNNs have achieved great success in various high-dimensional, nonlinear regression problems arising in computer vision, speech recognition, natural language processing, etc. While traditional statistical learning uses relatively simple models built on handcrafted input features, deep networks learn powerful task-specific features from raw, low-level input. We develop a network that maps signal features $x$ to transformed features $\phi(x)$ that accurately capture perceptual distances.

\begin{figure}[!t]
\centering
\includegraphics[scale=0.28]{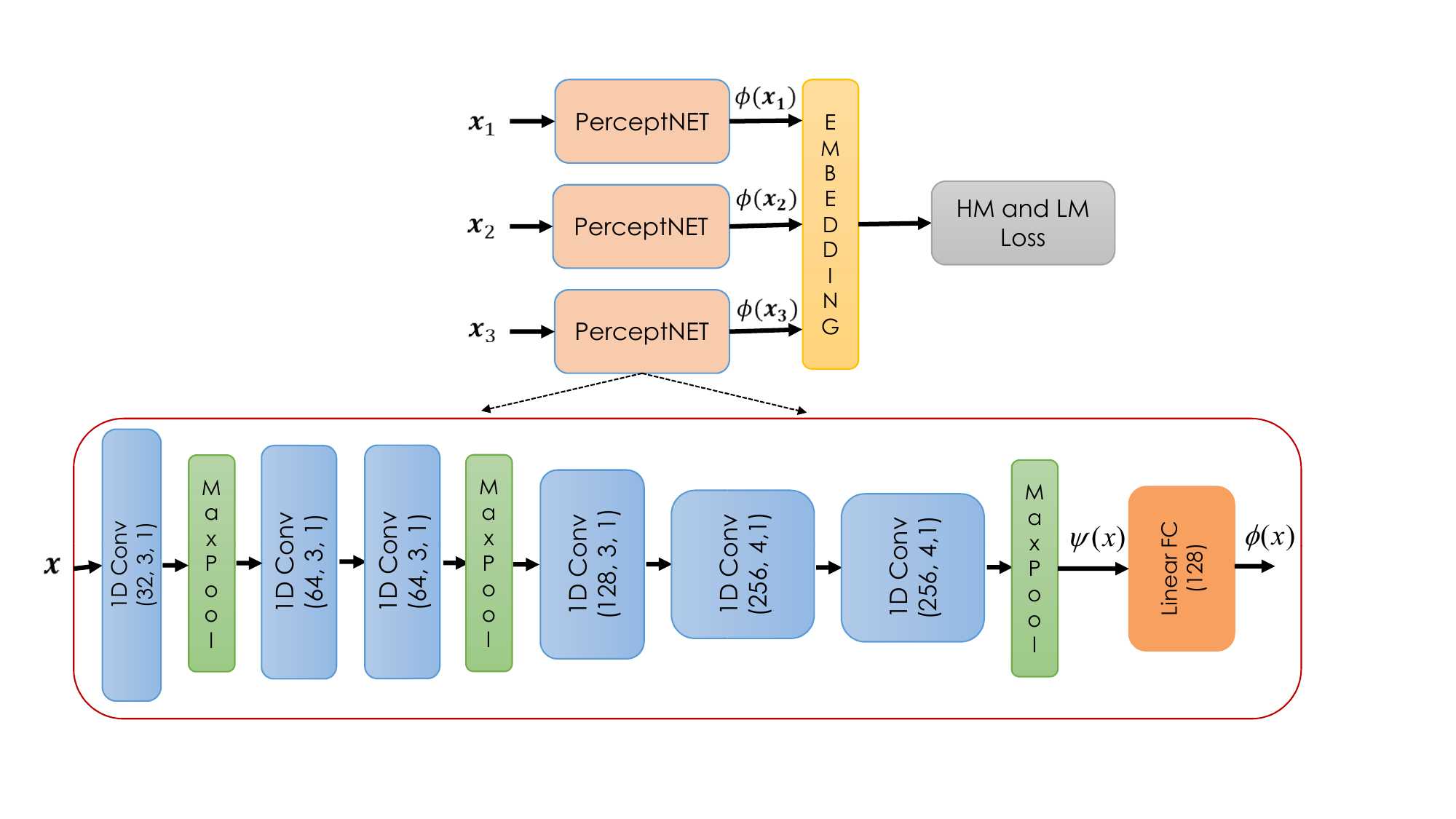}
\vspace{-10mm}
\caption{PerceptNet architecture. The network has six 1D convolutional layers, three pooling layers, and a linear fully-connected layer.}
\vspace{-3mm}
\label{architecture}
\end{figure}

Our model, dubbed PerceptNet, takes as input a spectral representation $x$ of the acceleration trace of a haptic texture (details below). The network consists of a series of convolutional layers, interspersed with max-pooling layers for downsampling, and ending with a fully-connected layer with {\em linear} activation and zero bias that outputs a 128-dimensional feature vector $\phi(x)$ (architectural details in Figure~\ref{architecture}). Hence, the network can be thought of as a {\em fully convolutional} portion $\psi(x)$, followed by multiplication with a matrix $W$ (the linear fully-connected layer): $\phi(x) = W^T \psi(x)$. The perceptual distance between signals $x$ and $y$ is then:
\begin{equation}
\begin{alignedat}{2}
  d_\phi(x, y) & ~=~ && \| \phi(x) - \phi(y) \| ~=~ \| W^T \psi(x) - W^T \psi(y) \| \\
               & ~=~ && \sqrt{ \left( \psi(x) - \psi(y) \right)^T W W^T \left( \psi(x) - \psi(y) \right) } \\
               & ~=~ && \sqrt{ \left( \psi(x) - \psi(y) \right)^T M \left( \psi(x) - \psi(y) \right) },
\end{alignedat}
\end{equation}
where $M$ is symmetric and positive semi-definite. In other words, the network effectively learns a Mahalanobis distance on the fully-convolutional embedding, which is a useful alternative visualization motivating the network design.

PerceptNet is trained with a novel loss that tries to satisfy both high- and low-margin constraints. Learning an optimal model in the presence of two kinds of triplets with opposed objectives requires carefully selecting the loss function. We must maximize the distance margin \mbox{$d_\phi(x_i, x_k) - d_\phi(x_i, x_j)$} for a high-margin triplet \mbox{$(x_i, x_j, x_k) \in H$}, while simultaneously minimizing the margin \mbox{$|d_\phi(x_i, x_k) - d_\phi(x_i, x_j)|$} for a low-margin triplet $\in L$. Following standard formulations of the triplet loss~\cite{FaceNet, LMNN, BoostMetric}, we modify the margins slightly to express them as differences of squared distances. This effectively weights long baseline triplets a bit more to correct gross errors in the learned manifold (although the loss without squaring gave comparable results in our experiments). The overall loss is \mbox{$E = E_H + E_L$}, where
\begin{equation} \label{loss}
\begin{alignedat}{1}
& E_H = \sum_{c \in H} \exp \left( -\rho(c) \right), \quad E_L = \sum_{c \in L} \left( 1 - \exp \left( -| \rho(c) | \right) \right) \\
& \text{and~} \rho\left( (x_i, x_j, x_k) \right) = d^2_\phi(x_i, x_k) - d^2_\phi(x_i, x_j)
\end{alignedat}
\end{equation}
The exponential allows training to focus on hard-to-satisfy triplets. The constant $1$ in $E_L$ is included only to visualize both losses (per triplet) in $[0, 1]$, and may be omitted. For each triplet, we pass its three signals through three copies of the network (with shared weights) and suitably penalize the distance margin, depending on whether the triplet is high- or low-margin. The gradient of the loss, obtained by back-propagating the training error, is used to iteratively update the network weights and improve the model with the Adam optimizer \cite{Adam} with a learning rate of 0.001. The model is implemented in PyTorch and trained for 1000 epochs with a batch size of 128. Given a trained model, we estimate the testing margin $\xi_\phi$, which will be used to classify test triplets as high- or low-margin, by minimizing $|f_H - f_L|$, where $f_H$ and $f_L$ are the fractions of correctly classified high- and low-margin {\em training} triplets respectively.

\subsection{CQFB Spectral Features}
\label{cqfb}

While deep networks largely eliminate the need to handcraft input features, they can still be helped by simple transformations of the input that may be hard for them to learn. In our case, we found that applying an initial spectral transform to the haptic signal, compactly summarizing periodic patterns in the data, led to better metric learning.

We use DFT321~\cite{DFT321} to find the spectrum (magnitudes) of 3-axis acceleration data. We divide the spectrum into 32 bins, increasing geometrically in size by a factor of 1.8 ~\cite{psychophysics1}, using a Gaussian filter of standard deviation 20 for some overlap between adjacent bins. We call this the constant $Q$-factor filter bank (CQFB) feature vector. Note that since bins are ordered coherently by increasing frequency, convolution is a meaningful operation on this vector.

\section{Experiments} \label{experiments}

We evaluate the effectiveness of our model through several experiments. First, we study synthetic datasets to validate our  algorithm in known linear and non-linear metric spaces. Then, we study a more challenging real-world haptic texture dataset \cite{TUM}. The performance of each model is evaluated by the fraction of satisfied triplet constraints (including, for high-margin triplets, predicting the correct ordering) in a held-out test set: the {\em triplet generalization accuracy} (TGA).

\subsection{Experiments on Synthetic Dataset}

\begin{figure}[!t]
\centering
\includegraphics[scale=0.265]{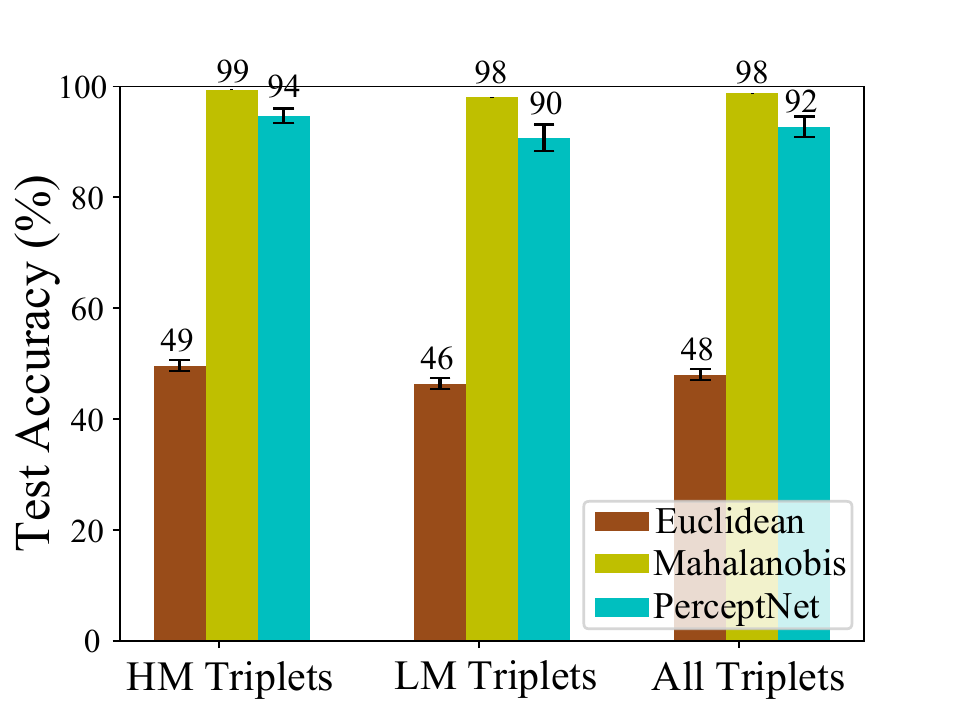}\hspace{-1mm}
\includegraphics[scale=0.265]{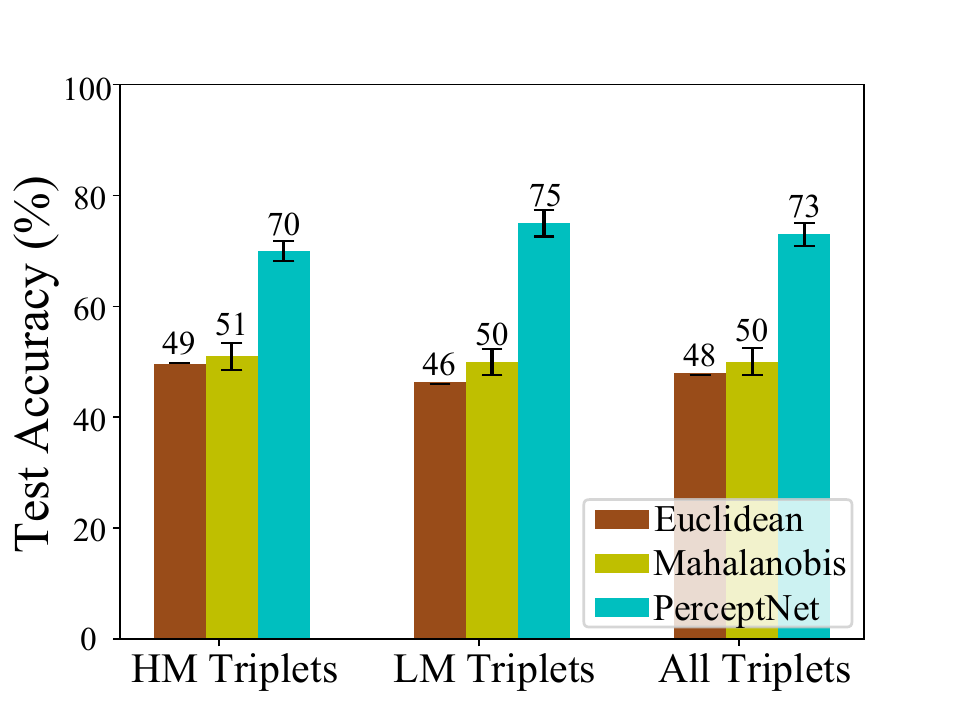}
\vspace{-5mm}
\caption{Triplet generalization accuracy (TGA) of PerceptNet compared to baselines, on synthetic datasets using linear Mahalanobis metric {\em (left)}; and non-linear elliptic Cayley-Klein metric {\em (right)}. Error bars represent standard deviation across 5 folds.}
\vspace{-3mm}
\label{fig:syn_bar}
\end{figure}

We first test our method with synthetic linear and nonlinear ground-truth metrics. For each metric, we draw 100 sample signals from an 8D standard normal distribution. We fix a random training threshold to distinguish between high- and low-margin triplets, according to the metric. We then randomly sample $10,000$ high-margin and $10,000$ low-margin training triplets. Each high-margin triplet is ordered appropriately. Similarly, we generate $20,000$ (10K + 10K) test triplets. The metrics we consider are:
\begin{enumerate}
  \item A linear Mahalanobis metric defined by a random PSD matrix $M$, and
  \item A nonlinear elliptic Cayley-Klein distance~\cite{Cayley} defined by a random invertible symmetric matrix $\Psi$.
\end{enumerate}
We repeat each experiment 5-fold for statistical significance, comparing the performance of PerceptNet with the natural Euclidean distance in CQFB space, and a learned Mahalanobis distance. The results are shown in Figure~\ref{fig:syn_bar}. With a ground-truth Mahalanobis metric, the learned Mahalanobis metric expectedly gives near-perfect accuracy, followed closely by PerceptNet. However, with nonlinear ground-truth, the limitations of linear modeling become clear and PerceptNet pulls significantly ahead of the other two.

\subsection{Experiments on Haptic Texture Dataset} \label{exp:texture_exp}

The TUM texture dataset~\cite{TUM} has acceleration signals recorded by freehand-tracing a sensing stylus over surface materials from 108 classes (metals, papers, grass, etc), with 10 signals per class. The authors also measured the perceptual similarity of each pair of classes by asking 30 subjects to distinguish between signals drawn from them. Instead of directly crowdsourcing triplet comparisons in an explicit new study, we reuse the data from the TUM study to construct training and testing triplets. This allows us to efficiently sample high- and low-margin triplets, using a standard dataset that has been used in a wide range of prior work. We define the ground-truth perceptual distance $d^*(x, y)$ between two signals $x, y$ as the fraction of subjects who could distinguish between the corresponding classes, normalized to $[0, 1]$. We also define the ground-truth perceptual margin $\xi^*$ as 10\% of the maximum margin over all possible triplets of signals.

Based on this dataset, we perform 3 experiments, ordered here from easiest to hardest. In each case, we repeat the experiment 5-fold, over 5 random train/test splits, and aggregate results over the folds for statistical robustness.

\begin{figure*}[!t]
\centering
\begin{tabular}{cccc}
\includegraphics[scale=0.33]{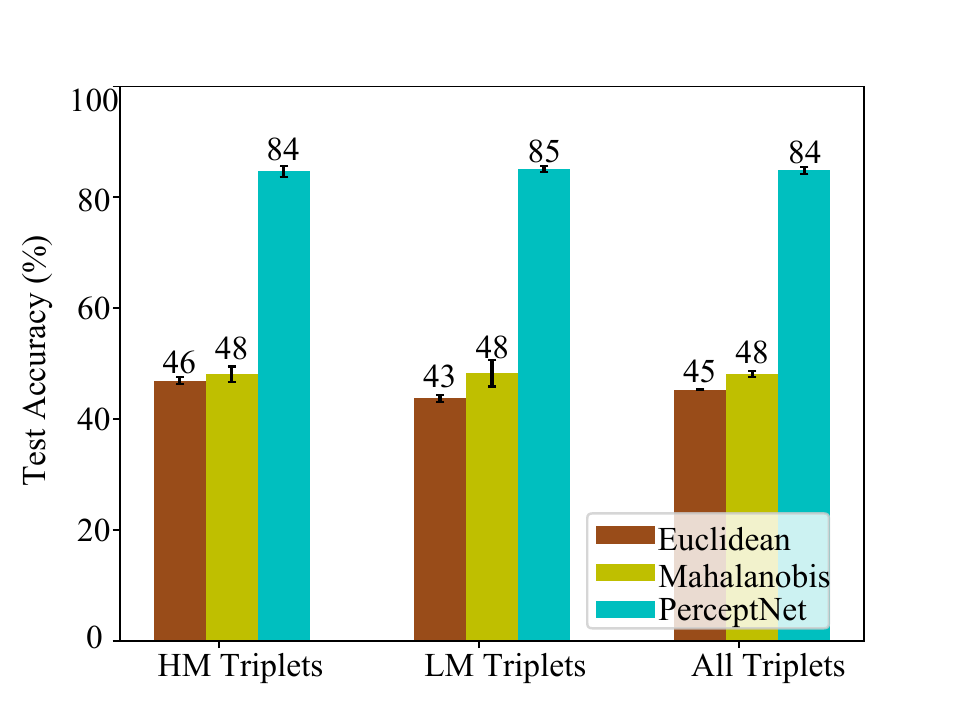} &
\includegraphics[scale=0.33]{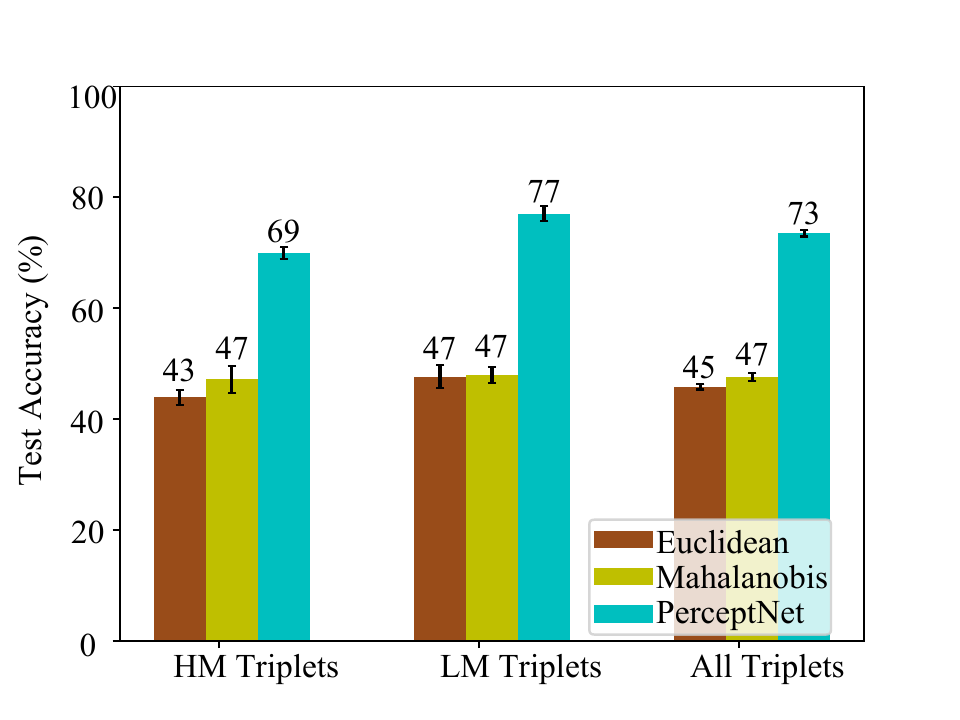} &
\includegraphics[scale=0.33]{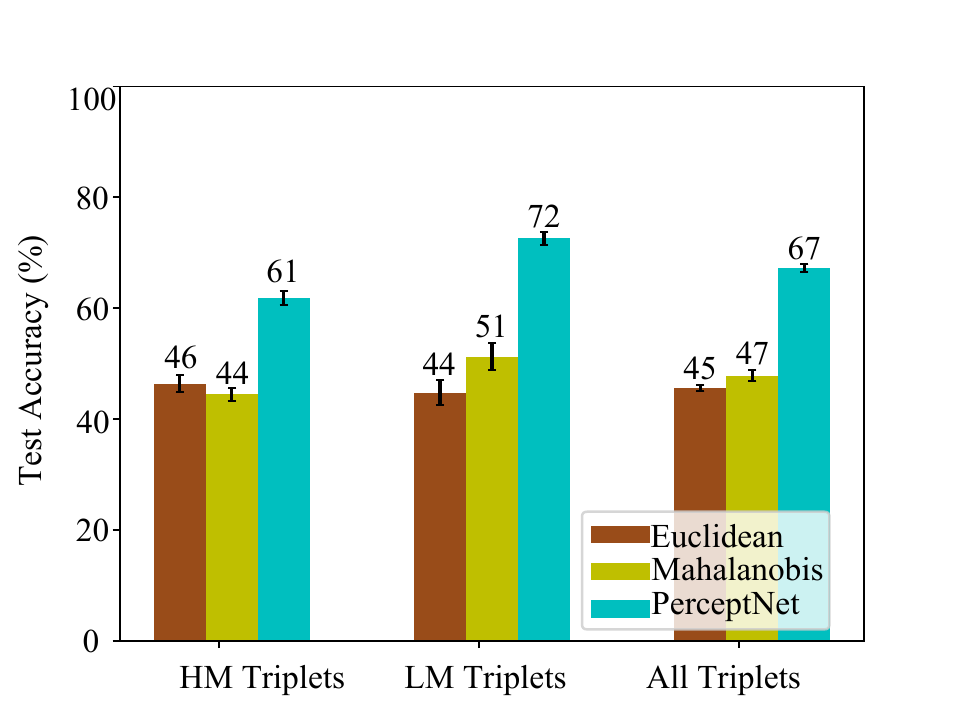}\vspace{-2mm}\\
\includegraphics[scale=0.33]{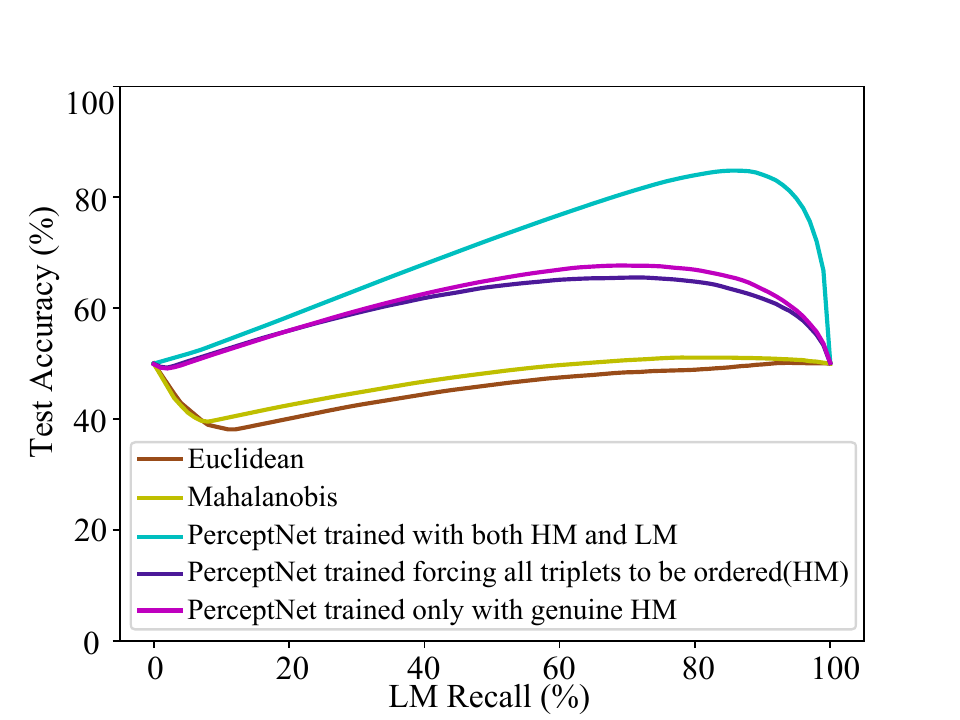} &
\includegraphics[scale=0.33]{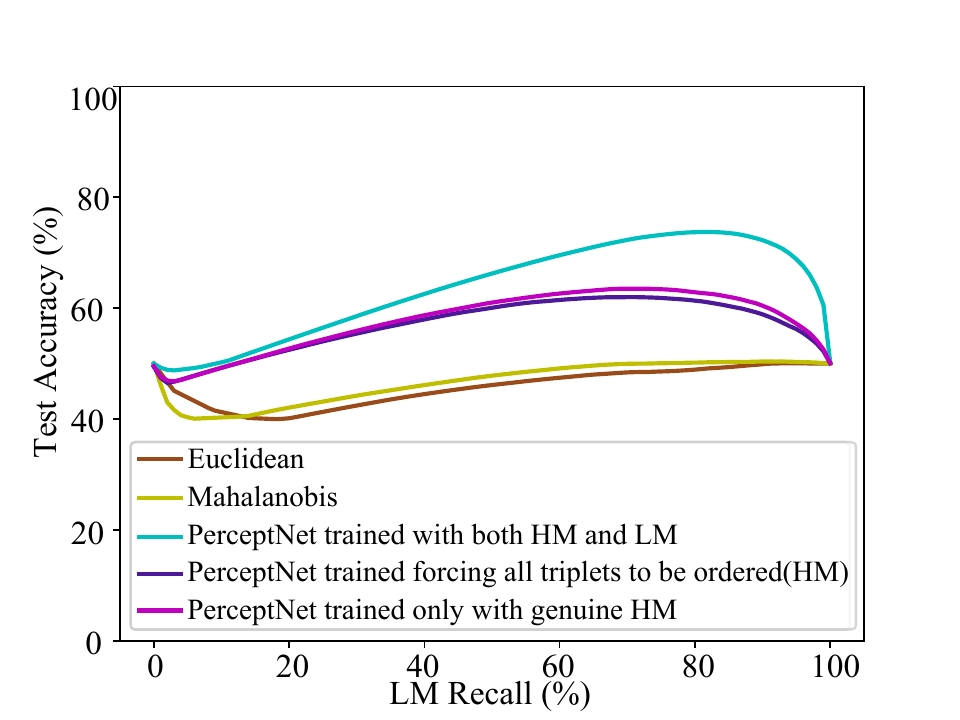} &
\includegraphics[scale=0.33]{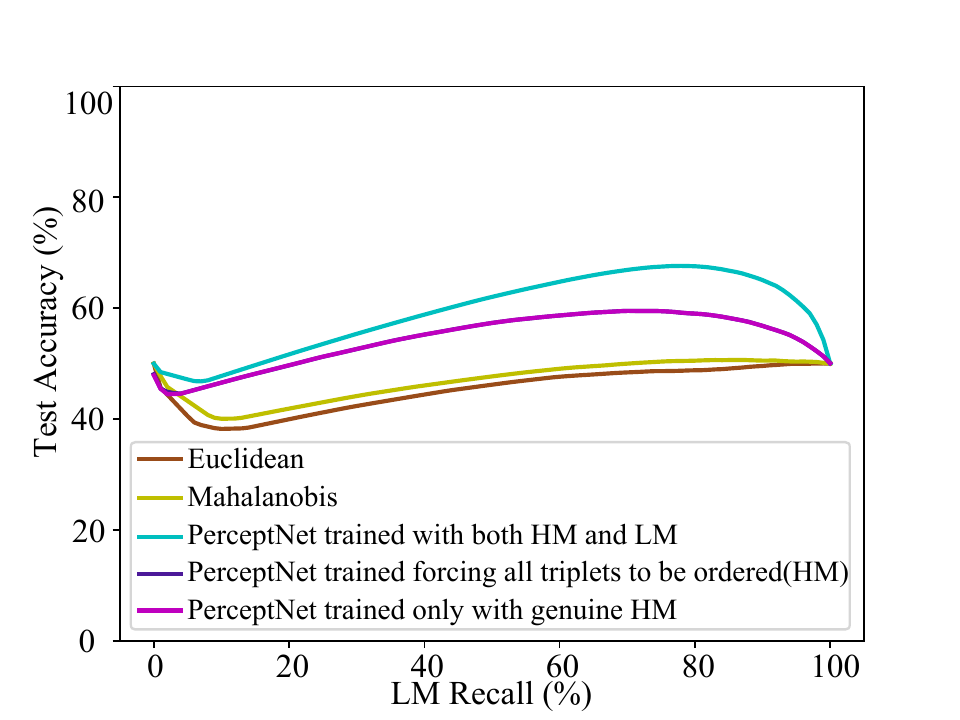} \\
\textbf{(a)} Held-out Triplets  & \textbf{(b)} Held-out Samples & \textbf{(c)} Held-out Classes  \\
\end{tabular}
\vspace{1mm}
\caption{Triplet generalization accuracy (TGA) for all three experiments of Section \ref{exp:texture_exp}-(a,b,c). We show both the optimal accuracy estimated using the learned threshold {\em (top)}, as well as the accuracy over the full range of thresholds {\em (bottom)}. To normalize different threshold ranges across different metric models, we map recall of low-margin triplets to the horizontal axis. In the bottom plots we also plot the accuracy of the same network architecture, trained without unorderable (low-margin) triplets. Note that in the absence of low-margin training triplets, the method cannot learn an optimal HM-vs-LM threshold, and hence does not have entries in the bar graphs above: the performance gap at any given threshold can be judged from the bottom plots. Our model significantly outperforms the baselines in all three experiments.}
\vspace{-3mm}
\label{fig:accuracy_comp}
\end{figure*}

\paragraph{Held-Out Triplets} In this experiment, we randomly sample $20,000$ triplets of signals for training, and $20,000$ different triplets for testing. We independently sample the complete collections of high- and low-margin triplets, determined by $d^*$ and $\xi^*$ (above), to ensure each set is an even mix of $10,000$ high-margin and $10,000$ low-margin triplets for balanced training and testing. Each high-margin triplet is ordered appropriately. Any signal may appear in both training and testing triplets: thus, in this experiment, we only study how well our method generalizes to unseen comparisons, not to unseen signals. Given training and testing sets, we compare the performance of a trained PerceptNet to the natural Euclidean metric and a learned Mahalanobis metric. The accuracies of the three methods are compared in Figure~\ref{fig:accuracy_comp}(a)(top): PerceptNet (84\%) significantly outperforms the alternatives. To factor out discrepancies caused by inaccurate estimation of the testing margin $\xi_\phi$ for any metric, we also plot the performance of the three methods over all possible margins (from 0 to 100\% recall of low-margin triplets) in Figure~\ref{fig:accuracy_comp}(a)(bottom). Again, PerceptNet is better than the alternatives over the entire range.

\begin{figure}[!t]
\centering
\includegraphics[width=0.49\linewidth]{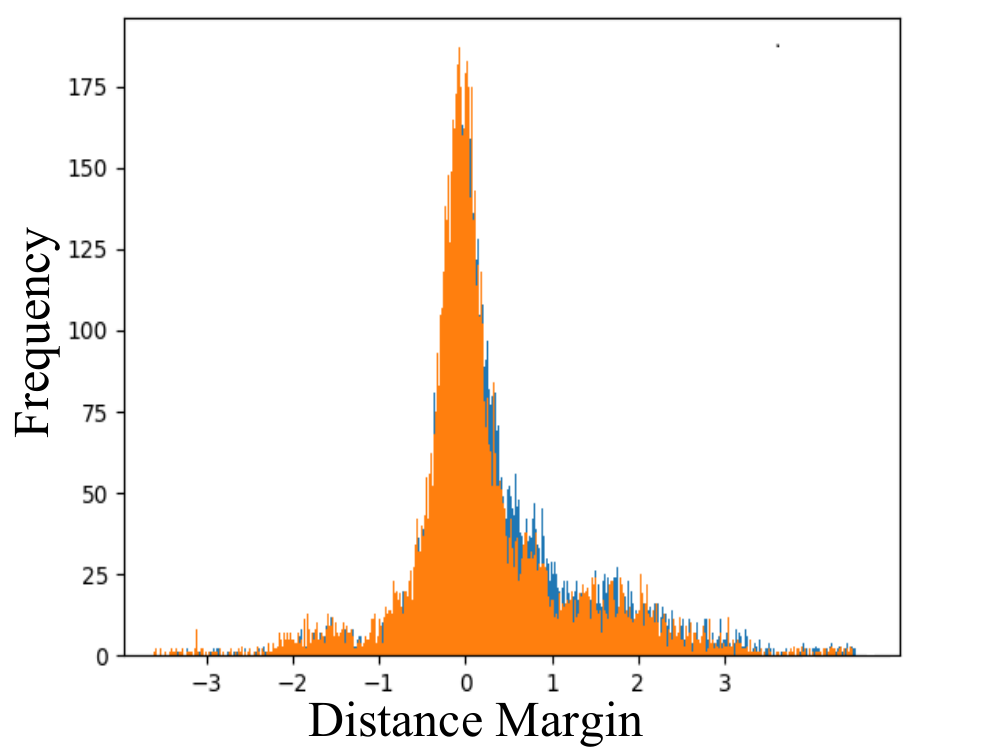}
\includegraphics[width=0.49\linewidth]{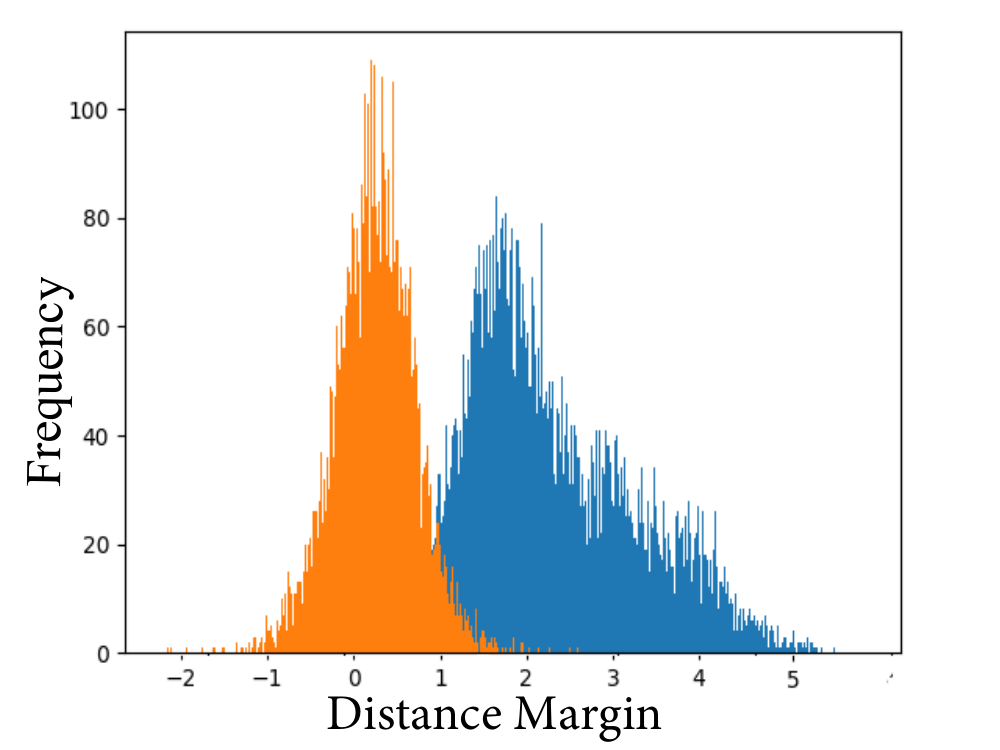}
\vspace{-5mm}
\caption{Distribution of learned high-margin (blue) and low-margin (orange) triplet margins generated from texture data. {\em (Left)} In Mahalanobis space. {\em (Right)} In PerceptNet space.\vspace{-3mm}}
\label{fig:hist}
\end{figure}

To gain insight into the embedding induced by PerceptNet, we plot a histogram of test triplet margins in Figure~\ref{fig:hist}. In the learned embedding space, high- and low-margin triplets have well-separated distributions. In contrast, a learned Mahalanobis metric fails to effectively separate the two types of triplets, leading to poorer predictions.

\paragraph{Held-Out Samples} Next, we randomly select 8 samples from each class for training, and hold out the remaining 2 samples for testing. We generate training and testing triplets as in the previous case, but with the additional constraint that training triplets must only use training samples, and testing triplets only use testing samples. This is a harder case since the method must generalize to unseen samples, albeit from known classes. The results are shown in Figure~\ref{fig:accuracy_comp}(b). The accuracy of PerceptNet (73\%) reduces in this harder case, but it is still distinctly better than the baselines. 

\paragraph{Held-Out Classes} Finally, we hold out {\em all} samples from a random $20\%$ of the 108 classes for testing. This is extremely challenging since the method must generalize to materials it has never seen before, and the training and testing distributions are quite different. As expected, performance drops further, but PerceptNet still generalizes much better (67\%, Figure~\ref{fig:accuracy_comp}(c)).\par
We also observed that in all three experiments, training PerceptNet with CQFB features (TGA 84, 73 and 67\%) performs better than the original raw features (78, 64, 53\%). This is a slightly unfair comparison since we simply widened the kernel sizes to adapt PerceptNet to the much higher-dimensional raw data. A more complex, hand-tuned network may be able to achieve comparable results with raw input also. However, CQFB enables a simpler, more efficient architecture with less manual effort.

We can make some high-level observations from the trends in these results. In all three experiments, the performance of Euclidean and Mahalanobis metrics remains $\sim$50\%, showing that these linear models are insufficient to represent the complex nuances of perceptual similarity. Also, as problem complexity increases (Figures~\ref{fig:accuracy_comp} and \ref{fig:accuracy_comp} (a) to (c)), performance on low-margin triplets is a little better than on high-margin triplets. We believe this is so because correctly classifying a high-margin triplet requires both large separation {\em as well as} correct ordering of the signals, i.e. identifying which signal is nearer and which is further from the base signal. In contrast, a low-margin triplet requires only that the separation of the base signal from the other two signals be small.

In the rest of this section we present further experiments to validate important aspects of our method. Each is performed in the ``held-out samples'' scenario above: \mbox{\ref{exp:texture_exp}-(b)}.

\subsubsection{Importance of low-margin triplets}
A primary feature of our method is that we train with both inequality (high-margin) and equality (low-margin) constraints. We validate this choice by comparing it to two other models with the same network architecture. The first generates training triplets with \mbox{$\xi^* = 0$}, forcing each triplet to be ordered (i.e. high-margin) even when participants found it difficult to determine such an ordering. Normally, triplets would be treated as unorderable/equidistant (i.e low-margin) to handle precisely this ambiguity. Hence, the data is noisy. The second model is trained with only high-margin triplets, completely ignoring the low-margin triplets (for fairness, we pick double the usual number of HM triplets, i.e. 20,000, to match the standard training set size). Hence, it cannot leverage equality information from the latter. All models are tested on the same test set containing a mix of high- and low-margin triplets. The bottom plots of Figure~\ref{fig:accuracy_comp} show that PerceptNet trained with both types of triplets more accurately models the underlying metric and outperforms the otherwise identical models trained with only high-margin triplets.

\begin{figure}[!b]
\centering
\vspace{-5mm}
\includegraphics[scale=0.35]{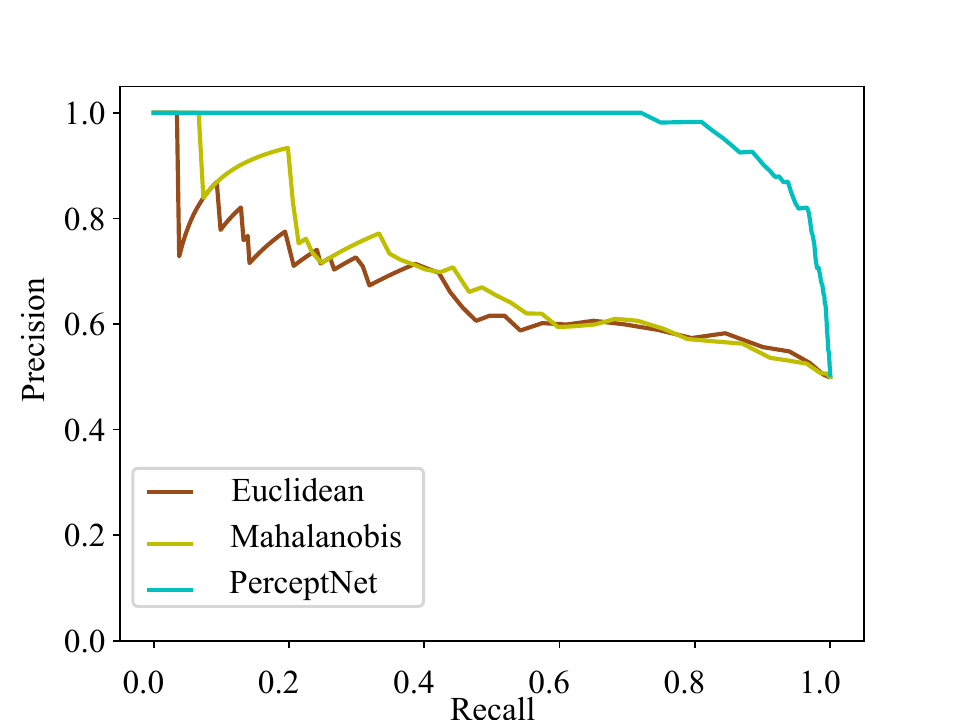}
\vspace{-2mm}
\caption{Precision-recall plot for classifying distinguishable and indistinguishable pairs of signals.}
\vspace{-2mm}
\label{fig:PR_pair}
\end{figure}

\subsubsection{Pairwise distinguishability}
\begin{figure}[!t]
\centering
\includegraphics[scale=0.28]{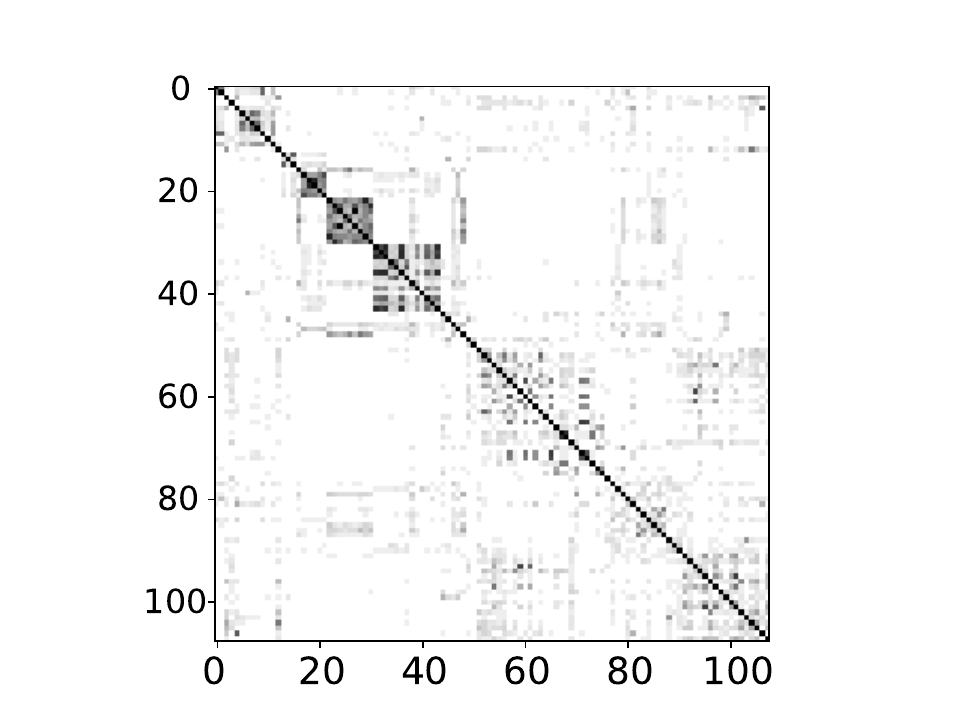}\hspace{-8mm}
\includegraphics[scale=0.28]{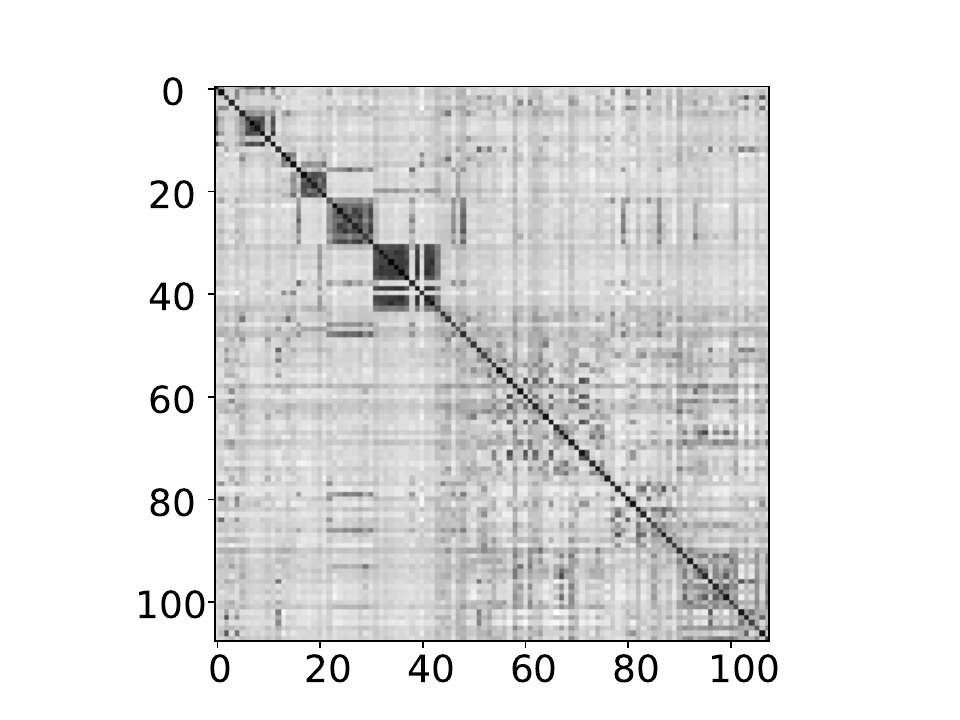}
\vspace{-3mm}
\caption{Perceptual similarity matrices of 108 material classes: the ground-truth confusion matrix from~\cite{TUM} {\em (left)} and PerceptNet distances {\em (right)}. White indicates low and black high similarity. Since PerceptNet is trained only on non-numerical triplets and need not have a linear relationship with the numerical ground-truth confusion values, the two matrices do not have the same normalization and are not identically mapped to the greyscale range. However, the trends in the two matrices are near-identical, indicating the accuracy of the model in ranking perceptual similarity.\vspace{-4mm}}
\label{fig:conf_mat}
\end{figure}

The original perceptual similarity ground-truth~\cite{TUM} provides pairwise measurements, which we exploit to generate triplets. Applications like vocabulary design also use pairwise comparisons \cite{maclean2003perceptual}. Hence, we verify if our learned metric can accurately predict distinguishable and indistinguishable signal pairs. We consider a pair distinguishable if $\geq 50\%$ subjects can tell them apart. We train PerceptNet with triplets as usual, but test on pairs of held-out signals, using a testing threshold to classify them as distinguishable or not based on their predicted separation. Figure~\ref{fig:PR_pair} shows a precision/recall plot over the full range of thresholds. Our method is significantly more accurate (AUC = 0.97) than alternatives (AUC = 0.69, 0.66). We also show that the pairwise similarity matrix on test signals, where each entry is the normalized average pairwise distance, captures very similar trends as the GT confusion matrix (Figure~\ref{fig:conf_mat}).

\subsubsection{Dependence on training set size}
\begin{figure}[!t]
\centering
\includegraphics[scale=0.35]{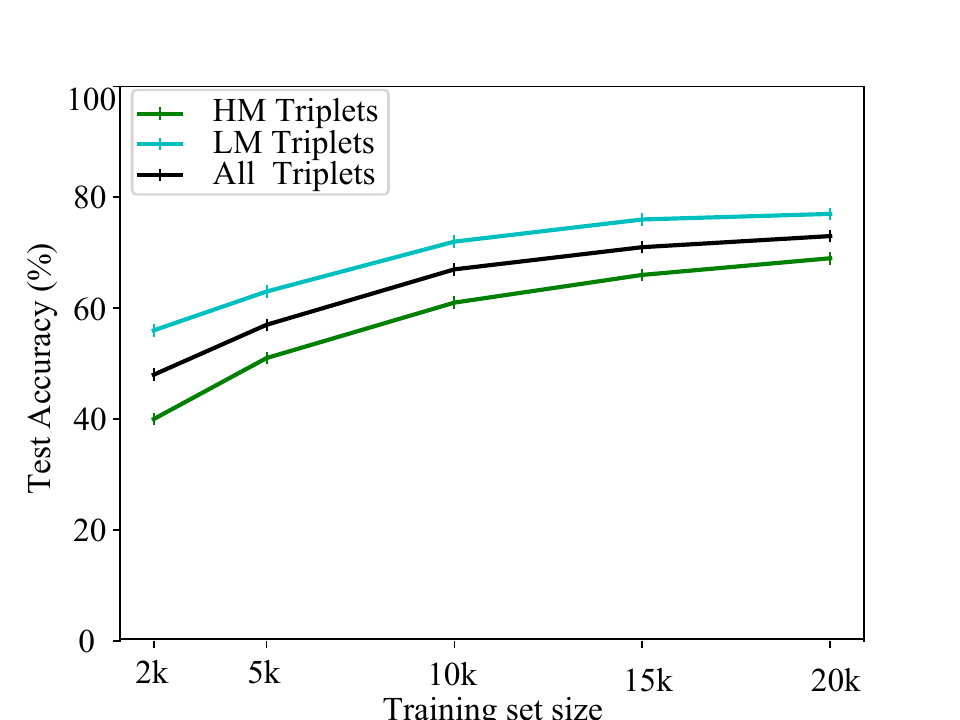}
\vspace{-3mm}
\caption{Triplet generalization accuracy of PerceptNet for different training set sizes. Accuracy increases monotonically, but with decreasing benefits for larger sizes.}
\vspace{-2mm}
\label{fig:training_set_size}
\end{figure}

Our last experiment evaluates how our model generalizes with the number of training triplets. Given a ``full'' training set of 20K triplets, we randomly sample subsets of different sizes and train a model on each such subset. As Figure~\ref{fig:training_set_size} shows, accuracy increases proportionally with the size of the training set.

\section{Conclusion}
In this work, we presented a deep metric learning approach for modeling the perceptual similarity of haptic signals. The model provides an end-to-end training framework for learning discriminative perceptual features from non-numerical triplet comparisons. Most significantly, we demonstrate the value of incorporating unorderable/equidistant signals into the training process, in order to better encode the nuances and limitations of human perception. Through extensive experiments, we show the our method's advantages over traditional metrics and its capacity to generalize to new data. We also find that the compact initial encoding of high-dimensional, highly-correlated, haptic acceleration traces as CQFB spectral features improves the accuracy of the learned model. In the future, we would like to model more complex nuances of human perception (e.g. just-noticeable difference), generalize to wider ranges of novel signals, and incrementally train the model in an online fashion.

\smallskip
\noindent {\bf Acknowledgments.} We thank Amit More for useful discussions which improved the paper.

\bibliographystyle{abbrv}
\bibliography{reference}

\end{document}